# GAME INFORMATION SYSTEM

Spits Warnars

Department of Computing and Mathematics, Manchester Metropolitan University,
Manchester M15GD, United Kingdom
s.warnars@mmu.ac.uk

## ABSTRACT

*In this Information system age many organizations consider information system as their weapon to compete or gain competitive advantage or give the best services for non profit organizations. Game Information System as combining Information System and game is breakthrough to achieve organizations' performance. The Game Information System will run the Information System with game and how game can be implemented to run the Information System. Game is not only for fun and entertainment, but will be a challenge to combine fun and entertainment with Information System. The Challenge to run the information system with entertainment, deliver the entertainment with information system all at once. Game information system can be implemented in many sectors as like the information system itself but in difference's view. A view of game which people can joy and happy and do their transaction as a fun things.*

## 1. INTRODUCTION

There are many kinds of Information system and have been implemented in many sector. They are come in all shapes and sizes and interweave one and others, no matter what management level use that information system. Indeed there are some information system have been designed as management level's need. They are such as:

1) Geographic Information system
2) Sale Information System
3) New Student Information system
4) Hospital Information system
5) Executive Information System
6) Etc

They use information system for their naming although there are some which do not use information system for their naming like Decision Support System, Expert System, Transactional Processing system and etc[1]. Whatever the names, use information system name or not, they are part of information system. They can be called sub information system which can make collaboration each others as one information system to deliver the best information system services.

Each of Information sector has their own specific character as a definition for each of sector. Example if we talk about Geographic Information system then will be associated with map. If we talk about hospital information system then will be associated with healthcare industry and if we talk about Sale information system then will be associated with selling. The same as if we talk about Game Information System then will be associated with game.

## 2. PROBLEM DEFINITION

Information system is an arrangement of people, data, processes, and information technology that interact to collect, process, store, and provide as output the information needed to support an organization [4]. There are many information systems which become sub information system that will collaborate one and others in one information system. They are :

1) TPS (Transactional Processing System)
2) SCM (Supply Chain Management)
3) CRM (Customer Relationship Management)
4) OLTP (Online Transactional Processing)
5) ES (Expert System)
6) EIS (Executive Information System)
7) MIS (Management Information System)
8) DW (Data Warehouse)
9) BI (Business Intelligence)
10) OLAP (Online Analytical Processing)
11) DSS (Decision Support System)

In the implementation the information system has been created as management level's needed. For example like TPS, OLTP, CRM, and SCM are designed for low level management to capture data and MIS, DW, OLAP, Expert System and DSS are designed for middle management, while EIS is designed for high level management. Although for some information system are designed for all management level like SCM,CRM, OLAP, DW, Expert System, and DSS. Figure 1 shows this type of information system and the level management allocation.

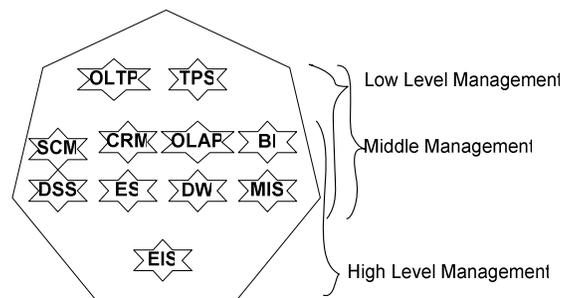

Figure 1. Type of Information System

A game is a structured or semi-structured activity, usually undertaken for enjoyment and sometimes also used as an educational tool [3,10]. Game is different from multimedia and is not just using multimedia elements like text, sound, picture, graphic art, animation and video [6] as shown in figure 2. Game is not just only present a real time or interactive entertainment like multimedia but there will be a play, fun and entertainment. Game uses all multimedia elements to enhance the game performance.

In multimedia people can interact with multimedia system or other people by teleconference and sometimes is present with 3D technology [8] but with game people can make the decision and train their strategy. By multimedia expert game has been known as multimedia at home or Virtual Reality [6].

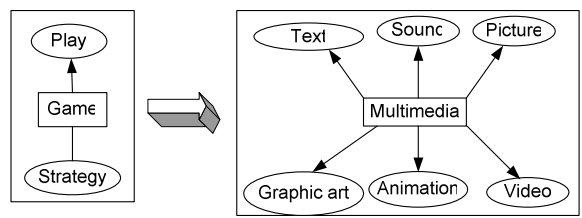

Figure 2. Game use multimedia elements for enhancement

There are thought that game is identical with violence, bad habits and has many negative thinking in our society. There are many violent games which have circulated and become favourites for young people[12]. There are no many love games which teach to love, care and patient. Game is always identical with competition and war. Parents always worry when their kids play too much game and sometimes they behave like the character in their game, beat their brothers, sisters or friends and even their dog [11]. Parents always worry when their children rebel and yell on their face, run out for their education and even they become lazy for their life and only waste their time in front of their game machine without any enhancement for their skills.

There are many bad characters in games which influence their character, and many these young boys want to become like the hero on their game. As a result for protection there are many violence games has been banned from sale and there will force to carry cigarette-style health warnings under proposals to protect children from unsuitable digital material.

In one hand for young people it is their rights to explore their appetite for playing game, their rights to explore their imagination, their rights to learn and do their experiment and their passion to know new things. In the right hands the young man's appetite will be handled. Our duty as a parents and researchers try to change this bad paradigm and to transform their big appetite for game as a positive trigger to enhance their skills. Our job is to soften and to change the bad paradigm about game and make the game as a tool for humans to explore their imagination. Sometimes game in the right using will become a good tool for humans to make a better life for this world.

Game information system as a new paradigm will be as way out for problem explaining before, and some questions below will be discussed in this paper as guidance to find the solutions. The questions are:

1) What is Game Information System?
2) What the differentiation between Game Information System with other Information systems?
3) What current people say about Game Information System?
4) How can make the Game Information System will be different with other Information System?
5) How to combine between fun and entertainment with information system?

## 3. GAME INFORMATION SYSTEM ARCHITECTURE

For standard game information system there are 3 optional for player to make transaction with the Information system as shown in figure 3. Player can chooses by how they interact with the information system !

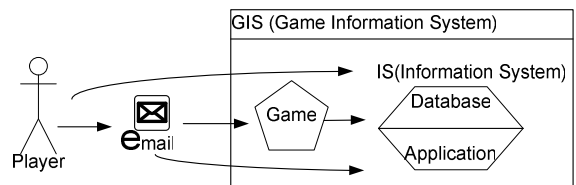

Figure 3. Standard game information system architecture

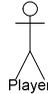

Figure 4.   Player

Player is the actor who get involve with the information system and player, which represent in figure 4. Player consist of :

1)  External player like customer, supplier, and student.
2)  Internal player like employee from high to low level management.

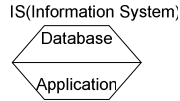

Figure 5.   IS framework

Information System will be implemented with information technology support. The architecture for information system will depend on each of organization, the organization's need. The organization can develops with management level's need, with technology like Data Warehouse, Decision support system, OLAP, OLTP and etc.

Application for information system can be built as standalone application or web application which will depends on the transaction, budget, knowledge and management's support. For standalone application can be built with Visual Basic, Java, dot net and etc. For web application can be built with Server programming like ASP (Active Server Pages), PHP (Personal Home Pages), JSP (Java Server Pages) and etc.  Of course for web application will need client programming like HTML (Hypertext Markup Language), Javascript or Jscript, CSS (Code Style Sheet) and etc.

The Information System will need database as shown in figure 5 and database as their data saving and for the performance the company can arranges how they implement their database architecture. The same with database the application will be put in the right proportion to make the good architecture implementation with 2 tier, 3 tier or more tier.

Choosing what kind of software application and database which will be used will depend on organization itself. Best Performance, easy to use and maintain, achievable budget can be used as indicator to make software justification. Also with hardware's choosing will depend on how the organization's budget and organization's vision and mission [9] to use information technology.

The three optional for player to make transaction as shown in figure 3 are :

1)  Traditional transaction
    Player transact with Information System as usual as a real human who is present on business site. Application can be developed become standalone application if only need one computer and can be extended with networking by LAN (local Area Network) or MAN (Metropolitan Area Network). Developing information system application for traditional transaction can be extended with web application. Figure 6 shows this traditional transaction.

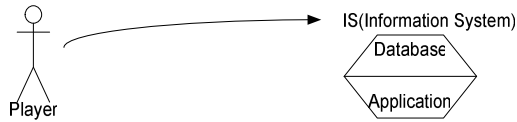

Figure 6. Traditional transaction

2) Email transaction

Player transact with Information System by email communication and even with teleconference as advancement. Player do not need present on business site. Email as a internet identity for player on ecommerce or ebusiness. Information system application must be present as web application for user convenience. Figure 7 shows this email transaction.

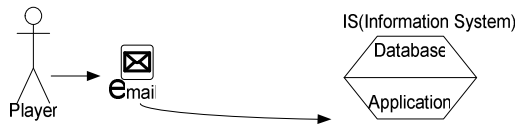

Figure 7. Email transaction

3) Game transaction

Player transact with Information System by Game framework. Email still to be needed to secure for each player as internet identity to represent the player itself. Email is needed as a ticket to make a sure for the player's interest with game. Information system application must be present as web application for user convenience. For game framework can be developed with game application like Java3D, ForGL and etc. Web application can be used to build this game framework. Figure 8 shows this game transaction.

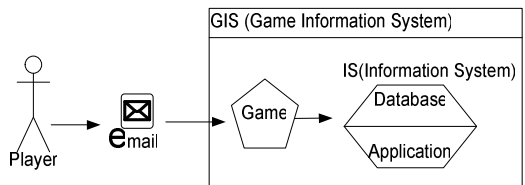

Figure 8. Game transaction

In reality world, there are many kind of people with different kinds of interest. There are people who are interested with traditional ways and there are people who like with new things or even with game. For some people using technology make them become slower and will increase their time and money for hardware and software. In order to fulfill all the external player's satisfaction then all the three of player's optional will be good if can be applied in one. Player can be freely to choose by what they want to access the information system. The right investment and future's thinking from high level management as the best valuable for company to go forward.

By making the three optional in one then in the middle of the transaction player can switch the way they interact with the information system. The player can switch from game transaction to email transaction or even to the traditional transaction. By easy to switch for player increase the easy for player specifically for external player to use for the system to handle every bad case situation like internet disconnection, disaster, fast services and etc.

Achieving best performance for game information system then we can extend from standard game information system to extended game information system. Game Information System will be more powerful if can be mixed with Virtual Application as shown in figure 9 and virtual information system as a part and the heart of Virtual Information System [5]. Virtual reality world will joins with game and make collaboration with information system to services their player.

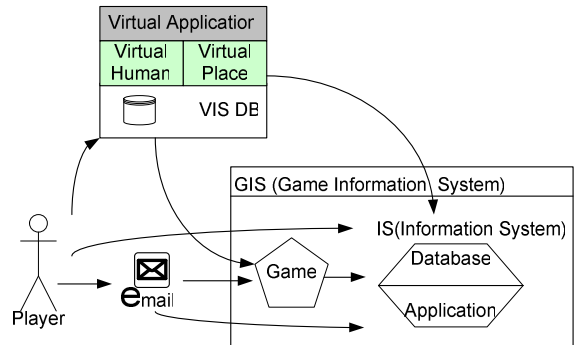

Figure 9.  Extended game information system architecture

Thus the three optional from standard game information system will transform become five optional in extended game information system.

In virtual application player can talk with other player as an avatar [14]. Every player's activities or conversation or any knowledge will be saved on VIS DB database as mining or knowledge[15]. Virtual application will be run as web application where people can access anywhere, anytime. Player only visits some limit area as their need on virtual place which will be controlled by unreal avatar.

Virtual application as shown in figure 9 will be divided as virtual human and virtual place.

Virtual Human is a part of virtual application to manage an avatar both real and unreal personal. All players both internal and external no need to present themself physically in real working place area, but their presence will be known by other players in virtual application. Virtual human will be needed by remote user who is not physically located on working area and will be needed by mobile user who location is constantly changing [7]. Figure 10 shows the transformation from real human to virtual human.

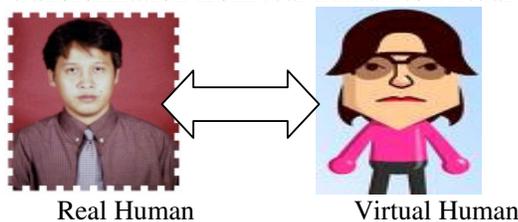

Real Human                Virtual Human
Figure 10. Real Human and Virtual Human

Virtual Place is a part of virtual application to manage real working place area. Virtual place will be limit by physical working area. Virtual place just only show up what the reality in real working place area and become the prohibition to show in virtual place something that never exist in real working place area. On the other hand we can not build a new building or change the room position without to make changing in virtual place and every changing in reality working place area must be recorded in virtual place. Everything look physically on working place area will be shown up in virtual place and also everything look virtually in virtual place is a mapping from reality working place area. Figure 11 shows the transformation real working place into virtual place and their connectivity.

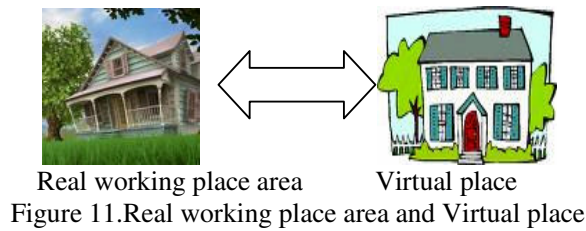

Real working place area          Virtual place

Figure 11.Real working place area and Virtual place

As an explanation before where standard game information system architecture in figure 3 will be more powerful when mixed with virtual application as shown in figure 9. The 2 next extended game information system player's optional are:

1) Virtual Game Transaction

Player transact with Information System by Virtual Game framework. Figure 12 shows how player connect with game information system by virtual application. Player can present as an avatar and present online on virtual place in virtual application and connect with game information system where having game and information system framework.

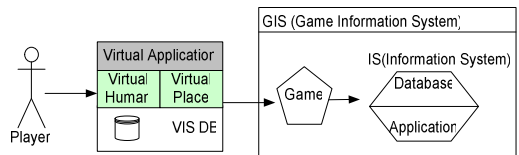

Figure 12.  Virtual Game transaction

2) Virtual Transaction

Player transact with Information System by Virtual information system, where virtual application will be combined with information system as virtual information system. There will need extended research for this virtual transaction as virtual information system. There is no game framework in this virtual transaction. Figure 13 shows how player connect with virtual information system where there are virtual application and information system.

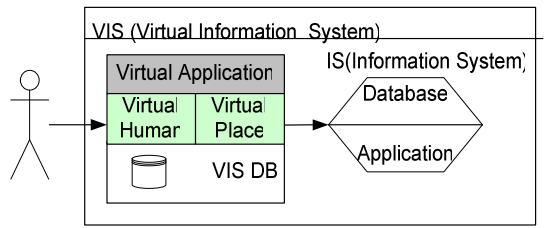

Figure  13.  Virtual transaction

In this extended game information system player will be presented with virtual world as enhancement for game information system for both of player's optional. The difference between virtual game transaction and virtual transaction is if we want to transact by virtually with information system with game atmosphere then we can choose virtual game transaction. On the other hand if we want to transact by virtually with information system without game atmosphere then we can choose virtual transaction.

By extending game information system as 5 player's optionals will extend the variance of the game information system. Player can be faced by many transaction's variation and every player will be satisfied by game information system as a purpose of information system itself.

## 4. GAME FRAMEWORK DESIGN

If game will be designed then the focus for the game must be determined. Game's Focus is determined on 4 focuses [17] as shown in figure 14, they are consist of :

1)  Play, can be played.
2)  Social, is there any interaction with other player
3)  Story, characters that follow very particular paths in their lives to create the right dramatic tension
4)  Simulation, there are simulated world.

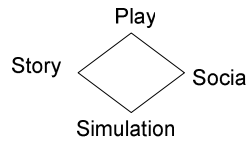

Figure 14. Focus for game design

Game's focus can be only one focus or more than one focus. These are the example of game's focus :

1)  Play, where the game can be played.
2)  Social, where the game can be played by more than one player.
3)  Story, where every company will choose each of their story or theme for their game according to company's vision and mission. For charming reason company can make the story more than one to decrease player's boring.
4)  Simulation, by making simulation world with virtual human and virtual place as part of virtual application.

If we choose standard game information system framework then we just implement play, social and story. But if we choose extended game information system by virtual application with virtual human and virtual place then simulation focus will be implemented either.

By making story, the game framework will be designed as the story will be showed as the organization's vision and mission. Player can choose in what age they will play and the more many ages optional will be presented the more player will be fun and happy. Figure 15 shows how the player can choose what kind of story they can choose.

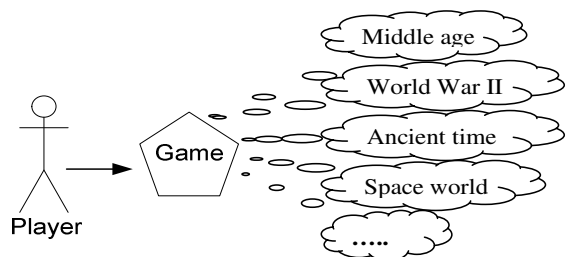

Figure 15. Actor can be in any age they wanted

After design the story then the player can choose what they want to be as shown in figure 16. The player optional acting will be the same road as the game's story. There is possibility where character not in line with the story and this feature may become opportunity and differentiate in game world. For example player can choose batman character to play in world's war II ages.

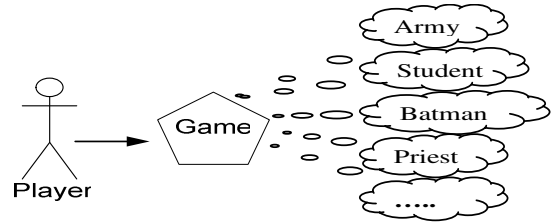

Figure 16. Actor can be any character they wanted

After declare the game focus then the company will decide what game types that will be implemented. Game's genre or type consist of [2] :

1) action
2) adventure
3) casual (board games, card games, and game show)
4) educational
5) RPGs (Role Playing Game)
6) Simulations
7) sports (including fighting games)
8) strategy
9) Other (puzzle games and toys).

Many games cross genre boundaries ("hybrids") or are truly several genres in one. Company can choose game's genre or multiple types. Some regulation will be needed to boost the player's interest like rewards.

Rewards can improve and extend the game play and, hence, provide a strong motivation for people to try to achieve the rewards [18,19]. Rewards will be done by collecting point and will be better if can be saved by player. In this Game rewards can be divided as 2 kinds of rewards like unreal rewards and real rewards as shown in figure 17. Unreal rewards is a reward where player can change their points with unreal rewards for player's satisfaction. For example they can change with weapon, clothe, food or science as the game's regulation. Real rewards is a rewards where player can change their points with real rewards like bonus, discount, souvenir or other promotions.

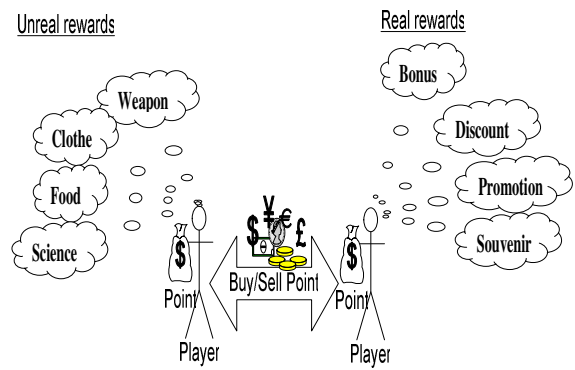

Figure 17. Player's rewards

Finally, player can buy and sell their points with other player and even they can get cash with many currency transactions. Even they can sell their property to other player.

## 5. BENCHMARKING

After we discussed game information system architecture and game framework design then we have answer some questions like what is game information system and of course we have assumed that game information system as the framework to present information system with game atmosphere where the player can joy and happy and do their transaction as a fun things. This game information system can be implemented to other information system.

Another questions like what the differentiation between game information system with other information system and how can makes the game information system will be different with other information system have been answered either. Design game information system architecture and game framework itself as differentiation with usual information system. Usual information have been designed as serious things must be done but game information system to mix game as a happy and joy ways to gain access information system. This answer has also answer the question how to combine between fun and entertainment with information system. Next we will answer what current people think about game information system.

Since 1990 game has been introduced for non entertainment purpose by some scholars. There is a term where look alike Game Information System that is called Serious Game. Serious Game is a software application developed with game technology and game design principles for a primary purpose other than pure entertainment [16]. They have been using for training, advertising, simulation or education in order to motivate and educate players.

The core segments of serious game are military, government, healthcare, corporate training, science, social change, first responder and education [13].

Serious game at www.socialimpactgames.com is categorized as
1) Education and learning games
2) Public Policy games
3) Political and Social games
4) Health and Wellness games
5) Business games
6) Military games
7) Advergames
8) Commercial (COTS) games

After looking one by one each of game at www.socialimpactgames.com then we assume that: all the games are serious game which is not fully serious game. All the games only give practicing, learning, teaching and simulation. Game is not fully get involve in real world as it is. So this is still we call unserious game as their motto at www.socialimpactgames.com is "The goal of this site is to catalog the growing number of video and computer games whose primary purpose is something other than to entertain". So the primary purpose is only to give practicing, learning, teaching and simulation and of course fun and entertainment as main of game itself.

Game information system as its ancestor can be implemented in many fields. For example :

1) New Student game Information System
   The New Students information system where can be accessed with game.
2) Sale Game Information System

    The Sale information system where can be accessed with game.

3)   Executive Game Information System

    The Executive Information System where can be accessed with game

4)   And etc

If we designed the game information system for new student information system then we will have new student game information system as combination between game information system and new student information system. Figure 18 shows this new student game information system as combination game information system and new student information system.

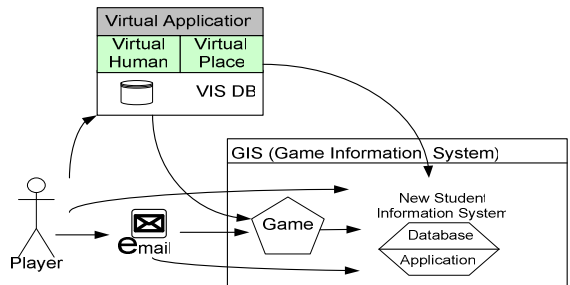

Figure 18. New student game information system

Also we can extend the new student for all university information system become university game information system as shown in figure 19. Particularly for the information system, there is possibility where the information system will be divided as OLAP (Online Analytical Processing) for high and middle management and OLTP (Online Transactional Processing) for low management.

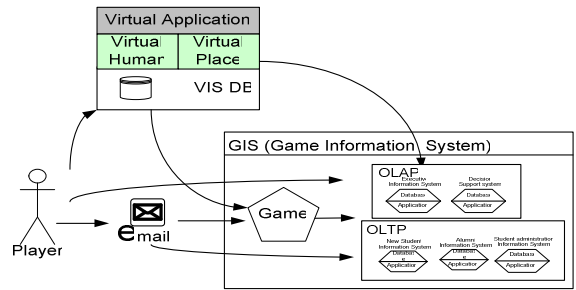

Figure 19. University game information system

There is possibility for sale game information system, the information system can be extended to handle SCM (Supply Chain management), CRM (Customer Relationship management), other OLTP (Online Transactional processing) and OLAP (Online Analytical processing). The extended sale game information system can be called corporate game information system as shown in figure 20.

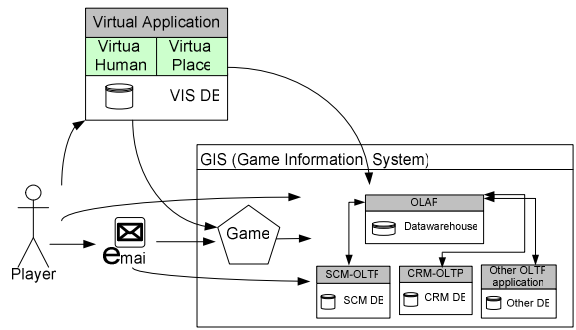

Figure 20. Corporate game information system

For example the story down here will explain what the new student game information system procedures.

1) As the ordinary new student information system, new candidate student must enroll to the system and also for new student game information system. New candidate student can play online from home or can access from the place which have been prepared by the university.
2) Game can be present as adventure type or casual type where the game framework will bring the new student candidate to enter their imagination by some multimedia presentation with sounds, graphic art, graphics 3D as shown in figure 21. The information with graphic presentation also can be showed for the convenience.

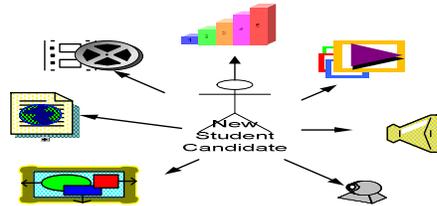

Figure 21. Multimedia presentation

3) New candidate student will be guided by some questions which can justify what level their skill and their knowledge about each of study program. In Traditional transaction new student candidate will face the test like English test, mathematical test, aptitude test, and Study program interesting test. The test result will decide what the new candidate student must do, how many money that they must pay for their installment fee and what the right study program for their skill and background. All the presentation will be in multimedia and there will be game concept such as play and do strategy.
4) Also new candidate student will be asked to choose and enter for some rooms as their science interest. Each of room will represent for each of study program and will be designed for skill and background purposes. Figure 22 shows the optional skill interest which can chosen by new candidate student. Off course not all new candidate student will like game, so new student game information system must prepare for its case as has discussed above as traditional or email transaction.

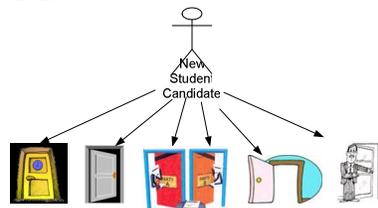

Figure 22. The doors of skill interest

5) Some little game which is casual game type like card game, board game can be present to catch the new candidate student skill and background.
6) After all the game's requirement have been fulfilled then the game can make the assessment what the right study program for new candidate student. The old knowledge which has been saved in data warehouse can be used as mining to sharpen the decision.

The advantages by mixing game and information system are:

1) As a differentiate, where we can use as one of the strategy to compete with our competitor
2) Reduce the people's stress in daily working

3)  Using the virtual application for game information system can be extended to virtual human community across to other game information system.

The disadvantages from game information system are:

1)  Will need investment extension for hardware and software in order to present the best services and easy to use system.
2)  Difficult for the first time specifically for people have no computer knowledge. Game is always identical with play and fun.
3)  People will focus to game rather than the information itself.
4)  The misuse of game information system will create bad imagination.

## 6. CONCLUSION

Because game is different with game information system then we must help the player with helping document or FAQ (frequently Asked question) to guide the lost player. There a little bit difficult for the first time, because our culture think that game will be associated with play and fun and more extreme as wasting time, violence, bad habits and others negative thinking .

We need time and help player as much as we can do, to train the player, and to change old concept about game which identical with negative habit become game as to do transaction with play and fun. Mirror game framework can be created to train the player and to prepare them in the real game for doing real transaction with play and fun.

The concept intermarriage between game and information system will bear a new career for expert who have knowledge in game and information system. The methodology for build the game information system as a framework can be deployed.

The implementation in one traditional information system like new student game information system will show the ability and the powerful include the disadvantages from game information system framework itself.

Another research can be deployed to keep search some knowledge which have connection with game information system like virtual information system, virtual application and etc.

## ACKNOWLEDGEMENTS

Thank you for DR. Edmond Prakash and DR. Nicholas Costen who give me inspiration for this research idea.

**Authors**

Bachelor's degree in Information System from University of Budi Luhur. Master's degree in Information Technology from University of Indonesia. A lecturer since 1995 at Information Technology faculty, University of Budi Luhur. Untemporal lecturer since Jan 2008 at Computer Science Department, University of Bina Nusantara. PhD Student for computer Science at Manchester Metropolitan University since Sept 2008.

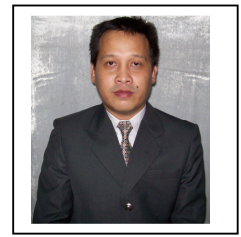